\documentclass[10pt,twocolumn,letterpaper]{article}

\usepackage{wacv}
\usepackage{booktabs}
\usepackage{multirow}
\usepackage{graphicx}

\usepackage{xcolor}
\usepackage{makecell}
\usepackage{mathtools}
\usepackage[accsupp]{axessibility} 

\definecolor{wacvblue}{rgb}{0.21,0.49,0.74}
\usepackage[pagebackref,breaklinks,colorlinks,allcolors=wacvblue]{hyperref}

\def\wacvPaperID{1060} 
\def\confName{WACV}
\def\confYear{2027}

\title{DOSE: Dominance-Guided Ordinal Severity Estimation \\from Sparse Relative Annotations}

\author{
Tushir Sahu \qquad Hideaki Hayashi \qquad Hajime Nagahara\\
The University of Osaka, Japan\\
{\tt\small tushir.sahu@is.ids.osaka-u.ac.jp, \{hayashi, nagahara\}@ids.osaka-u.ac.jp}
}

\begin{document}
\maketitle
\begin{abstract}

Medical severity estimation commonly relies on absolute grade annotations, which are fundamentally ordinal, yet existing methods often treat the grades as independent categories. Learning-to-rank methods, which can be trained using relative annotations based on severity comparisons of image pairs, can provide continuous severity rankings but cannot directly predict discrete severity grades. We propose \textbf{D}ominance-guided \textbf{O}rdinal \textbf{S}everity \textbf{E}stimation from Sparse Relative Annotations (DOSE), a framework for learning ordinal medical severity from sparse pairwise comparisons without using absolute annotations. DOSE aggregates the observed comparisons through a global comparison graph to estimate a dominance score for each image. Differences between dominance scores set cumulative margins in the embedding space, so that pairs with larger estimated severity differences are separated more strongly, and also generate pseudo-pairs that supervise a scalar severity head alongside the observed comparisons. The learned continuous severity axis is converted into discrete grades via label-free score calibration, with tie-anchored segmentation as the main calibration method. In the experiment, DOSE outperformed pairwise-ranking baselines in most evaluated settings on ACNE04, IDRiD, and LIMUC under different pair budgets. Code is available at \url{https://github.com/TushirSahu/DOSE}

\end{abstract}

    
\section{Introduction}
In the medical domain, automatic image-based disease severity estimation has become an important problem in computer-aided diagnosis, as it can help clinicians assess disease progression and make treatment decisions. Many deep learning methods have been widely applied to this problem in different medical imaging domains, including retinal imaging, endoscopy, and facial photography~\cite{cho2019automated,klang2021automated,lim2020automated,wang2021deep,wienholt2024ordinal}. There are several approaches to framing severity estimation using images; one method involves treating it simply as a standard classification problem, where images are assigned to one of several predefined, independent severity grades. Alternatively, to better account for the ordinal nature of disease severity, methods based on ordinal regression or ranking are also frequently considered. This is important because neighboring severity grades are usually more similar than distant grades, and an error between grades 1 and 2 should not be treated in the same way as an error between grades 1 and 4. 

Most severity estimation methods rely on absolute annotations, in which a medical expert assigns a single discrete grade to each training image. This annotation process can be difficult because disease appearance often changes continuously, while the clinical grading scale divides this into a small number of discrete categories. Images near a grade boundary can therefore be ambiguous even for experienced clinicians, leading to inter-observer variability. Kalpathy-Cramer et al.~\cite{kalpathy2016plus} reported poor agreement when experts assigned absolute retinopathy-of-prematurity categories. In contrast, the agreement was considerably stronger when the same images were compared by relative disease severity. Similar observations have been reported in biomedical image assessment~\cite{phelps2015pairwise,jang2022decreasing}.

Relative annotation provides an alternative supervision protocol in which an annotator labels an image pair according to which image is more severe, or marks the two as comparable. Such pairwise relations directly encode the ordered structure of disease severity without providing an absolute grade for either image. Previous work has shown that relative comparisons can be used to learn continuous disease severity rankings, and the annotation cost is lower than that of absolute annotation~\cite{kalpathy2016plus,kadota2024deep}. However, the number of possible image pairs grows quadratically with the dataset size, making it important to learn effectively from sparse comparisons.


Learning-to-rank is the natural approach to exploiting these relative annotations, but standard pairwise ranking has two limitations for severity grading. First, a pairwise ranking loss captures only the direction of severity: given an ordered pair $(x_i,x_j)$, a ranking loss encourages the score of the more severe image to be larger, but does not specify how much larger it should be. A mild--moderate pair and a mild--severe pair therefore impose essentially the same ordering constraint, although the latter should represent a larger ordinal separation. Second, the learned rank score is continuous, whereas many clinical practices require estimating discrete severity grades. 

In this work, we ask whether both ordinal separation and discrete severity grades can be recovered from sparse relative comparisons without absolute-grade supervision. We answer this by proposing \textbf{D}ominance-guided \textbf{O}rdinal \textbf{S}everity \textbf{E}stimation from Sparse Relative Annotations (DOSE). To address the first limitation, DOSE aggregates the sparse comparison graph into global dominance scores, whose differences provide a label-free estimate of ordinal separation. These differences determine the cumulative margins in the embedding space so that images with similar estimated severity remain closer together, while more distant images are separated more. A scalar severity head is trained in parallel on the observed comparisons together with confident dominance-derived pseudo-pairs. To address the second, DOSE converts the learned continuous severity axis into discrete grades via label-free score calibration, using annotated ties as anchors that keep equally severe images within a single grade. DOSE thus yields both a continuous severity ordering and discrete ordinal predictions without training a label-dependent classifier.

The main contributions of this work are: 
\begin{itemize}
\item We introduce a global dominance-based formulation that converts the sparse comparison structure into ordinal distance estimates, enabling cumulative multi-margin representation learning without absolute class labels. 
\item We propose dominance pseudo-pair supervision for the scalar severity head, which propagates transitive ordering information from the comparison graph to pairs that were not explicitly annotated.
\item We propose tie-anchored segmentation (Tie-seg), a label-free score-to-grade calibration, and compare it with GMM- and prior-based alternatives, as well as a known-prior oracle. \end{itemize}

\section{Related Work}
 
\subsection{Ordinal Learning for Medical Image Analysis}
 
Most clinical severity scales are ordered, but the losses typically used to fit them are not. The cross-entropy loss treats grade 0 and grade 3 as two arbitrary grades, so a model pays the same penalty whether it confuses two adjacent stages or two stages at opposite ends of the scale. For disease grading, that is the wrong consideration, since mistaking mild for moderate rarely alters treatment, whereas misinterpreting mild for severe frequently does. Early works adapted neural networks to ordered categories by redesigning the output layer~\cite{cheng2008neural}. Niu et al.~\cite{niu2016ordinal} proposed a $K$-way ordinal problem as a bank of binary subproblems solved jointly by a multiple-output CNN, and CORAL~\cite{cao2020rank} sharpened the idea by forcing those subtasks to share a weight vector, which makes the predicted cumulative probabilities monotonic by construction and removes the rank inconsistencies that affect independent binary heads. CORN~\cite{shi2023deep} relaxes the shared-weight constraint and models conditional rank probabilities instead, recovering consistency without tying the subtasks together. More recent work shifts the ordinal prior elsewhere entirely, with OrdinalCLIP~\cite{li2022ordinalclip} learning rank prompts in a language latent space, ensuring that the class prototypes emerge ordered by construction.

The same works are applied across different organs and modalities. Chen et al.~\cite{chen2019fully} grade knee osteoarthritis with an ordinal loss on Kellgren-Lawrence scores. Wu et al.~\cite{wu2019joint} jointly model acne severity and lesion count using label distribution learning, addressing overlap between neighboring grades. Polat et al.~\cite{polat2023improving} proposed a loss function that takes the order of classes and the distances of the classes into consideration for estimating the ulcerative severity. MORF~\cite{lei2022meta} couples a CNN with a differentiable regression forest inside a meta-learning loop. DOSE shares the same goal of preserving the ordered structure of severity grades.  

These ordinal learning methods rely on absolute grades to define their training objective. In contrast, DOSE learns the ordinal structure from sparse relative comparisons and does not use absolute grades as training targets.
 
\subsection{Contrastive and Ordinal Representation Learning}

Contrastive learning structures the embedding space by bringing related samples closer and separating dissimilar ones~\cite{chen2020simple,khosla2020supervised,he2020momentum,grill2020bootstrap,caron2021emerging,zbontar2021barlow,bardes2021vicreg}, while ordinal representation methods further consider the order or distance between targets. Rank-N Contrast~\cite{zha2023rank} preserves target order through rank-aware contrastive learning, and POE~\cite{li2021learning} models ordinal embeddings probabilistically to capture prediction uncertainty. SCOL~\cite{saleem2023scol} incorporates label-dependent ordinal distances into supervised contrastive learning, while SupReMix~\cite{wu2025supremix} uses ordinal information and pair hardness for contrastive regression. CLOC~\cite{pitawela2025cloc}, which is particularly related to our work, uses multiple margins to separate samples according to their ordinal classes.

However, these methods derive the ordinal structure from absolute labels or target values, whereas DOSE estimates it from dominance differences recovered from sparse relative comparisons.

\subsection{Pairwise Comparison and Dominance-Based Ranking}

Pairwise comparison provides a natural way to recover an ordering when absolute scores or full rankings are difficult to obtain. Classical models such as Bradley--Terry~\cite{bradley1952rank} estimate a latent strength for each item from observed comparison outcomes. RankNet~\cite{burges2005learning} extends this idea to neural models by learning a score function directly from a pairwise approach. Some simpler counting-based approaches aggregate the number of wins and losses associated with each item. Shah et al.~\cite{shah2018simple} showed that such estimators can be both robust and statistically effective under suitable assumptions. Graph-based ranking methods make use of the global structure of the comparison set. Rank Centrality~\cite{negahban2016rank} represents pairwise outcomes as a Markov chain and derives item scores from its stationary distribution. This allows evidence to propagate through indirect comparison paths rather than being limited to directly observed pairs. Heckel et al. ~\cite{heckel2018approximate} studied the number of comparisons required to recover an approximate ranking, while GNNRank~\cite{he2022gnnrank} learns global rankings from directed comparison graphs.

Also, relative supervision has been explored in computer vision. Parikh et al.~\cite{parikh2011relative} introduced relative attributes, where visual properties are learned from statements indicating that one category exhibits an attribute more strongly than another. Their work showed that pairwise relations can capture semantic structure that is difficult to express through independent categorical labels. Phelps et al.~\cite{phelps2015pairwise} found pairwise assessment to be more reliable than Likert-style rating for biomedical image quality. Jang et al. ~\cite{jang2022decreasing} further showed that human-in-the-loop pairwise sorting can reduce the annotation burden associated with expert assessment. Pairwise ranking objectives have also been used directly for medical grading, including colorectal cancer grading in pathology images ~\cite{le2021ranking}.

DOSE builds on the same pairwise supervision and global ranking principles but uses the resulting dominance scores differently. Rather than treating the graph-based rank as the final prediction, DOSE uses dominance as an intermediate source of supervision to estimate ordinal distances and generate pseudo-pairs.

\subsection{Discrete Grades Without Labels}

Learning an ordered representation and assigning a discrete grade are, although related, two separate problems. Supervised ordinal regression methods such as CORAL~\cite{cao2020rank} and CORN~\cite{shi2023deep} produce ordered predictions by using absolute ordinal labels during training. Pseudo-labeling and self-training provide another route by treating confident model predictions as temporary supervision ~\cite{sohn2020fixmatch}. In settings where no class labels are available, deep clustering methods instead recover discrete groups from learned features. Self-supervised approaches, like DeepCluster~\cite {caron2018deep}, alternate between clustering visual features and using the resulting assignments to update the representation. Invariant Information Clustering~\cite{ji2019invariant} learns assignments by maximizing mutual information between predictions for related views of an image. SeLa~\cite{asano2019self} jointly learns representations and self-labels through an optimal transport formulation with a balanced assignment constraint. SwAV~\cite{caron2020unsupervised} performs online clustering and learns by predicting the cluster assignment of one image view from another. SCAN~\cite{van2020scan} follows a two-stage formulation in which self-supervised representation learning is followed by neighbor-based semantic clustering. Together, these methods show that meaningful discrete partitions can be learned without absolute labels.

DOSE is related to these approaches in that it also obtains discrete
outputs without using absolute grades as training targets. However,
clustering methods produce unordered cluster identities, whereas the
severity axis in DOSE already has a low-to-high direction determined by the relative comparisons. 

\section{Proposed Method}
\label{sec:methodology}

\begin{figure*}
    \centering
    {\includegraphics[scale = 0.72]{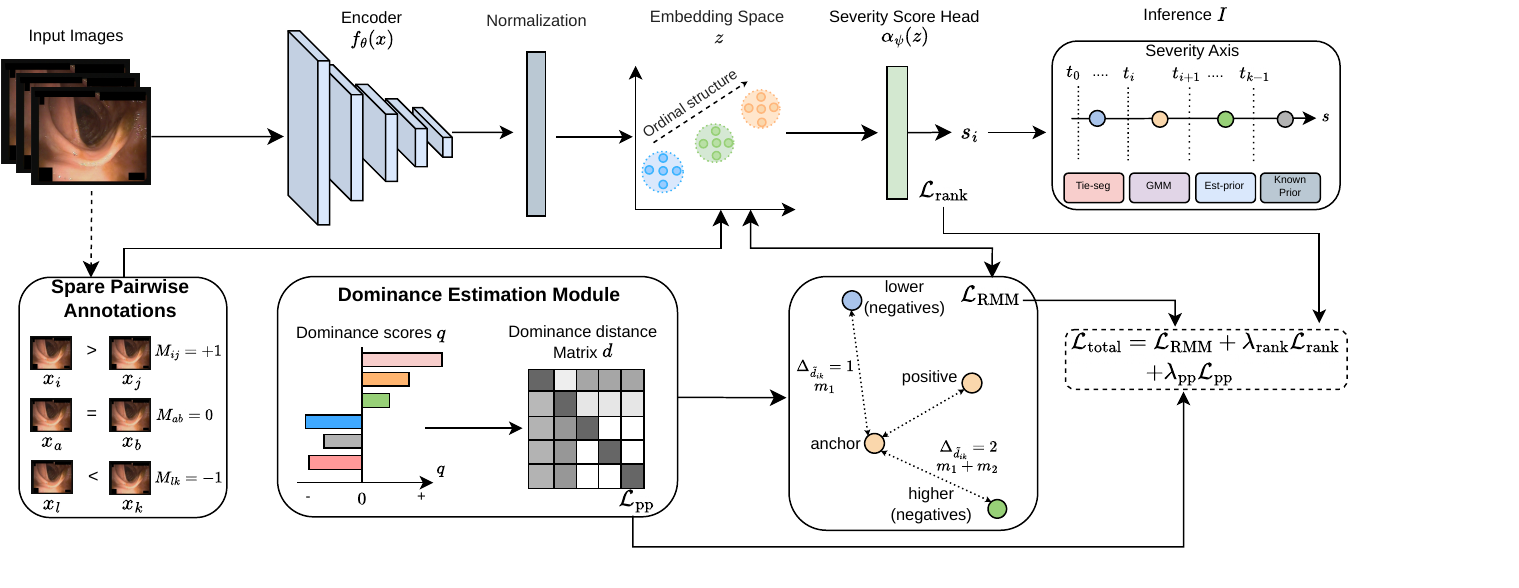}}
    \noindent
    \caption{The overall architecture of our proposed method DOSE}
    \label{fig:overview}
\end{figure*}
\noindent
\subsection{Overview}
Fig.~\ref{fig:overview} shows the overall framework of DOSE. We consider a set of training images $\mathcal{X}=\{\mathbf{x}_i\}_{i=1}^{N}$ with $K$ ordered severity levels, where supervision is available only through sparse pairwise
comparisons. For a pair $(\mathbf{x}_i,\mathbf{x}_j)$, we use $\mathbf{x}_i\succ \mathbf{x}_j$ when $\mathbf{x}_i$ is more severe than $\mathbf{x}_j$, $\mathbf{x}_i\prec \mathbf{x}_j$ when it is less severe, and $\mathbf{x}_i\sim \mathbf{x}_j$ when the two have comparable severity. The corresponding relative label is
\begin{equation}
\label{eq:rel}
r_{ij}=
\begin{cases}
+1, & \mathbf{x}_i\succ \mathbf{x}_j,\\
-1, & \mathbf{x}_i\prec \mathbf{x}_j,\\
0,  & \mathbf{x}_i\sim \mathbf{x}_j.
\end{cases}
\end{equation}

Our method consists of three main components. First, an image encoder produces an ordinal embedding $\mathbf{z}$, which is used both for representation learning and by a scalar head to estimate a severity score. Second, the observed comparisons are aggregated through the dominance estimation module to obtain a global dominance score for each training image. The dominance differences are converted into ordinal distance levels for the cumulative multi-margin loss and are also used to construct confident pseudo-pairs for the severity head. Finally, after training, the learned severity scores are converted into $K$ discrete grades using the score-calibration routes described in Sec.~\ref{subsec:calib}. 


\subsection{Encoder and Embedding Space}

Given an input image $\mathbf{x}$, the encoder $f_\theta$ produces a $d$-dimensional embedding,
\begin{equation}
\label{eq:heads}
\mathbf{z}
=
\frac{f_\theta(\mathbf{x})}
{\lVert f_\theta(\mathbf{x})\rVert_2},
\qquad
s=\alpha_\psi(\mathbf{z}),
\end{equation}
where $\mathbf{z}\in\mathbb{R}^{d}$ is the normalized embedding and $s\in\mathbb{R}$ is the continuous severity score. 

The embedding $\mathbf{z}$ is used by the ordinal representation loss, while the scalar head $\alpha_\psi$ maps the same representation onto a
one-dimensional severity axis. Here, the similarity between two embeddings is measured by
$S_{ij}=\frac{\mathbf z_i^{\top}\mathbf z_j}{\tau}$, where $\tau$ is the temperature parameter.

\subsection{From Pairwise Signs to Ordinal Distances}
\textbf{Sparse pairwise annotations.} Within a mini-batch of $B$ images, we store the observed comparisons in a sparse matrix
\begin{equation}
\label{eq:sparse}
M_{ij}=
\begin{cases}
r_{ij} & \text{if observed},\\
\emptyset & \text{otherwise}.
\end{cases}
\end{equation}
Under a realistic annotation budget, most entries remain empty, and unobserved pairs do not contribute to the training.

Now, with $B$ images in a batch, there are $\binom{B}{2}=B(B-1)/2$ possible pairwise comparisons. It is computationally expensive to compute all such pairs and introduces unnecessary redundancy. Instead, during each pairwise-encoding round, we partition the mini-batch into disjoint pairs and consider only $\left\lfloor\frac{B}{2}\right\rfloor$
comparisons. Thus, each image participates in at most one comparison in a sampling round, reducing the number of encoded comparisons from $\mathcal{O}(B^2)$ to $\mathcal{O}(B)$. However, this also limits the number of directly observed relationships. To address this, we created a dominance estimation module to mitigate this limitation.

\begin{figure*}
    \centering
    \makebox[\textwidth][c]
    {\includegraphics[scale = 0.8]{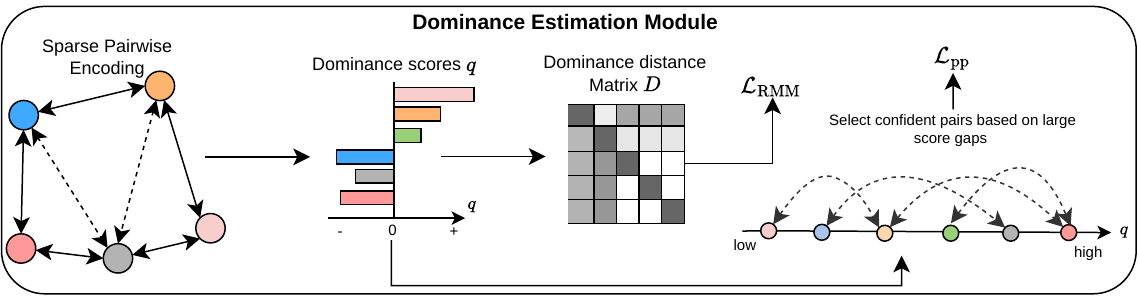}}
    \caption{The Dominance Estimation Module}
    \label{fig:dominance}
\end{figure*}

\textbf{Dominance Estimation Module.}
Fig.~\ref{fig:dominance} illustrates the overall architecture of the dominance estimation module.
A win-and-loss total calculated in a single batch $B$ is noisy and removes transitive evidence. Even though the two images were never directly compared, $\mathbf{x}_a$ should dominate $\mathbf{x}_c$ if $\mathbf{x}_a\!\succ\!\mathbf{x}_b$ and $\mathbf{x}_b\!\succ\!\mathbf{x}_c$. We recover this by estimating dominance once over the entire comparison graph rather than batch by batch. Let $A_{uv}$ count the observed comparisons in which $\mathbf{x}_u$ is judged more severe than $\mathbf{x}_v$. We build a row-stochastic random walk toward more severe images and obtain its distribution after $L$ iterations as a latent severity strength~\cite{negahban2016rank}. For observed non-tied pairs, the off-diagonal transitions are:
\begin{equation}
\label{eq:rc}
P_{{uv}} \propto \frac{A_{vu}}{A_{uv}+A_{vu}},\qquad
\boldsymbol{\pi}={(P^{\top})}^L\frac{\mathbf{1}}{N}.
\end{equation}
See supplementary for more details. 
An image that has won over other images many times in terms of severity should accumulate high probability mass, which is exactly the transitive signal that a batch-local tally misses. The resulting probabilities $\pi_i$ can be highly skewed, so we operate with their ranks instead. The dominance score $q_i$ is the rank of $\pi_i$ normalized to $[0,1]$, the fraction of images with smaller stationary mass,
\begin{equation}
\label{eq:q}
q_i=\frac{1}{N-1}\,\big|\{\,u : \pi_u<\pi_i\,\}\big|.
\end{equation}
Working with ranks keeps $q$ stable against the heavy-tailed values that $\pi$ can take on in an imbalanced graph. This score aggregates transitive evidence across the whole graph and is used to define ordinal distances and pseudo-pairs, but never the final prediction.
\noindent

\textbf{Dominance pseudo-pairs.} Only the pairs that are annotated are considered by $\mathcal{L}_\mathrm{rank}$. Since every image is already ordered by global dominance $q$, we use it to densify the pairwise signal at no additional annotation cost. A pseudo-pair with dominance label $\hat r_{ij}=\mathrm{sign}(q_i-q_j)$, treated as a fixed target, is formed by any batch pair with a consistent gap $|q_i-q_j|>\tau_{pp}$.
\begin{equation}
\label{eq:pp}
\begin{split}
\mathcal{L}_{\mathrm{pp}}=\frac{1}{|\tilde{\mathcal{P}}|}
\sum_{(i,j)\in\tilde{\mathcal{P}}}\big[\,m-\hat r_{ij}\,(s_i-s_j)\,\big]_{+},\\
\tilde{\mathcal{P}}=\{(i,j):|q_i-q_j|>\tau_{\mathrm{pp}}\}.
\end{split}
\end{equation}
This carries the transitive ordering over to pairs that were never directly annotated, again without any absolute label. 

\textbf{Ordinal distance.}
For an observed pair we measure the dominance gap $\hat d_{ij}=|q_i-q_j|$ and turn it into an ordinal-distance level in $\{1,\dots,K\!-\!1\}$. Rather than slice this gap on a fixed grid, we cut it at the empirical quantiles of the observed gaps, which keeps the levels balanced even when one grade dominates the data. Let $e_m$ denote the $\tfrac{m}{K-1}$ empirical quantile of $\{\hat d_\mathrm{uv}\}$ taken over all observed non-tied pairs, so the $K\!-\!2$ interior edges $\{e_m\}_{m=1}^{K-2}$ divide the gap distribution into $K\!-\!1$ equal-mass levels and every level holds a comparable share of pairs. The level of a pair is then the number of edges its gap exceeds,
\begin{equation}
\label{eq:dist}
\tilde d_{ij}=\mathrm{clamp} \!\Big (1+\!\sum_{m=1}^{K-2}\mathbb{I}\!\left[\hat d_{ij}>e_m\right],\,1,\,K\!-\!1\Big),
\end{equation}
with $\tilde d_{ij}=0$ for tied pairs, whereas $\tilde d_{ij}$ records how strongly two images should be separated, giving the label-free proxy for ordinal distance that the multi-margin loss consumes.

\subsection{Learning the Embedding and the Severity Axis}
\textbf{Dominance-aware multi-margin loss.}
For each anchor $\mathbf{x}_i$ we split the batch into positives $S_{\mathrm{pos}}(i)$ with $M_{ij}=0$ and ordinal negatives $S_{\mathrm{low}}(i)\cup S_{\mathrm{high}}(i)$ with $M_{ij}\neq0$. A single margin would push a grade-1 negative and a grade-3 negative equally far from the anchor, which is wrong for an ordinal target. We instead grow the margin with the ordinal distance. With learnable step margins $m_t=\mathrm{Softplus}(\theta_t)$, a label-free proxy ordinal distance $\tilde d_{ik}$ accumulates a cumulative margin $\Delta_{ik}=\sum_{t=1}^{\tilde d_{ik}} m_t$, so the required gap is monotonic in distance by construction. Over the triplet set $\mathcal{T}=\{(i,j,k): j\in S_{\mathrm{pos}}(i),\,k\in S_{\mathrm{low}}(i)\cup S_{\mathrm{high}}(i)\}$ the loss is
\begin{equation}
\label{eq:rmm}
\mathcal{L}_{\mathrm{RMM}}=\frac{1}{|\mathcal{T}|}\sum_{(i,j,k)\in\mathcal{T}}
\big[\,\Delta_{ik}-S_{ij}+S_{ik}\,\big]_{+}.
\end{equation}
We replace $\Delta_{ik}$ with a single shared margin to recover a single-margin variant that we report only as an ablation in Sec.~\ref{sec:ablation}.

\textbf{Severity ranking.}
In parallel, the scalar head is trained to assign higher scores to more severe images. With $\delta_{ij}=s_i-s_j$ over the non-tied set $\mathcal{P}_{\neq}=\{(i,j):r_{ij}\neq0\}$,
\begin{equation}
\label{eq:rank}
\mathcal{L}_{\mathrm{rank}}=\frac{1}{|\mathcal{P}_{\neq}|}\sum_{(i,j)\in\mathcal{P}_{\neq}}
\log\!\big(1+\exp(-r_{ij}\,\delta_{ij})\big).
\end{equation}

\begin{figure}
    \centering
    \makebox[\columnwidth][c]{\includegraphics[scale=0.62]{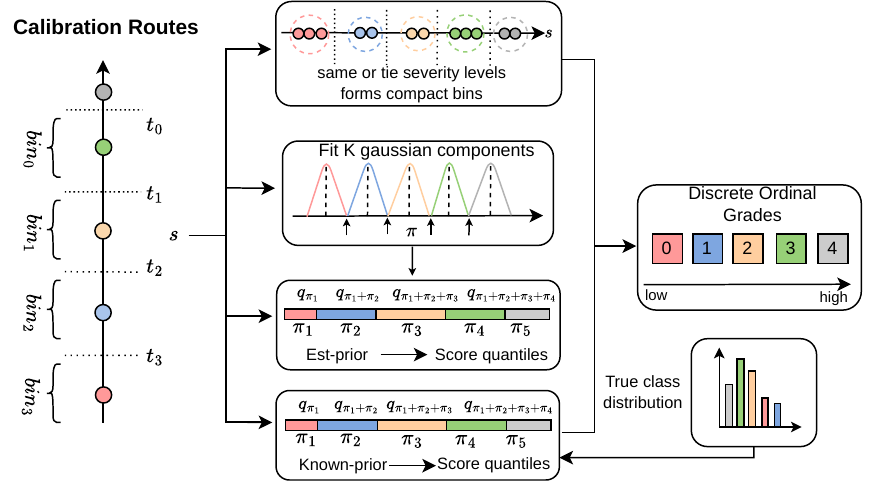}}
\caption{
Score-to-grade calibration routes. A learned scalar severity score is converted into discrete ordinal grades by placing $K-1$ cut-offs on the score axis. Tie-seg, GMM, and Est-prior are label-free routes, while Known-prior uses the true training
class proportions.}
    \label{fig:calibration}
\end{figure}
The three loss terms are optimized jointly to give us the total loss function as
\begin{equation}
\label{eq:total}
\mathcal{L}=\mathcal{L}_{\mathrm{RMM}}+\lambda_{\mathrm{rank}}\,\mathcal{L}_{\mathrm{rank}}+\lambda_{\mathrm{pp}}\,\mathcal{L}_{\mathrm{pp}}.
\end{equation}

\subsection{Reading Grades from the Severity Axis}
\label{subsec:calib}

After training, each image is represented by a continuous severity score $s_i=\alpha_\psi(\mathbf{z}_i)$. This score provides an ordinal ranking, but clinical grading requires a discrete severity level. We obtain a grade by placing $K-1$ ordered cut-offs $t_1<\cdots<t_{K-1}$ on the score axis and counting how many cut-offs the score exceeds:
\begin{equation}
\label{eq:digitize}
\hat y_i
=
\sum_{k=1}^{K-1}
\mathbb{I}[s_i>t_k].
\end{equation}

Since no absolute grades are available during training, these cut-offs cannot be learned from labeled validation data. Instead, they are estimated from the learned score
distribution and the relative global comparison structure. We study four cut-off placement routes as shown in Fig.~\ref{fig:calibration}. The first three are label-free. The fourth uses the true class prior and is reported only as known-prior (oracle). Let $s_{(1)}\le\cdots\le s_{(N)}$ denote the sorted training scores.

\textbf{Tie-seg (prior-free).}
We split the sorted scores into $K$ contiguous bins $B_1,\dots,B_K$ and pick the split that keeps each bin tight while keeping equally graded images together.
\begin{equation}
\label{eq:tieseg}
\min_{B_1,\dots,B_K}\! \sum_{b=1}^{K}\sum_{s\in B_b}\!(s-\mu_b)^2
\!+\!\lambda_{\mathrm{tie}}\!\!\!\!\sum_{(i,j)\in\mathcal{T}}\!\!\mathbb{I}[B(i)\neq B(j)],
\end{equation}
where $B(i)$ is the bin of image $i$, $\mu_b$ is the mean score of bin $B_b$, and $\mathcal{T}$ is the set of training tie pairs. The first term is the usual within-bin variance and favors compact bins. The second penalizes every tie the boundaries pull apart by $\lambda_{tie}$, so a pair annotated equally severe resists being split across a grade line. Ties act as soft anchors that hold the same-grade images within a single bin. Since the scores are one-dimensional and the cost is additive across bins, the optimal set of boundaries is obtained efficiently by sorting scores, without any threshold search. Each cut-off $t_k$ sits at the midpoint of the two adjacent scores that straddle a boundary. This route assumes neither a class prior nor a density model and applies to practical conditions.

\textbf{GMM.}
As a label-free parametric alternative, we fit a $K$ component one-dimensional Gaussian mixture to the training scores,
\begin{equation}
\label{eq:gmm}
p(s)=\sum_{k=1}^{K}\pi_k\,\mathcal{N}(s;\mu_k,\sigma_k^2).
\end{equation}
The mixture is fitted without using absolute labels. After fitting, we sort the components by their means, $\mu_1<\cdots<\mu_K$, so that the components follow the learned severity direction. We then place the cut-off between two adjacent grades at the midpoint of neighboring component means,
\begin{equation}
t_k=\frac{1}{2}(\mu_k+\mu_{k+1}), \qquad k=1,\dots,K-1 .
\end{equation}
This route assumes that each ordinal grade forms a distinct compact bin on the score axis. It is fully label-free, but it can become unstable when the score distribution is imbalanced or when adjacent grades overlap strongly.

\textbf{Est-prior.}
The GMM fit also gives mixture weights $\pi_k$, which provide a label-free
estimate of the class proportions. Instead of directly using the Gaussian component locations as boundaries, this route uses only the estimated prior. After sorting the mixture components by their means, we treat the reordered
weights as estimated grade frequencies and place cut-offs at the corresponding empirical score quantiles:
\begin{equation}
\label{eq:estprior}
t_k=Q\!\Big(\textstyle\sum_{l=1}^{k}\pi_l\Big),
\qquad k=1,\dots,K-1,
\end{equation}
where $Q(\cdot)$ is the empirical quantile function of the training scores. This separates the role of estimating the class prior from the role of assigning samples to Gaussian components. In other words, it tests whether an estimated label-free prior is sufficient to recover useful grade boundaries on the learned severity axis.

\textbf{Known-prior.}
Finally, we report an oracle quantile route that uses the true class proportions
$p_k$ from the training set:
\begin{equation}
\label{eq:knownprior}
t_k=Q\!\Big(\textstyle\sum_{l=1}^{k}p_l\Big).
\end{equation}
This route is not a label-free method. It is included only as an upper-bound reference, as it shows how well the learned severity axis can perform if the true class prior is available. 

\section{Experiments}

We evaluate DOSE on three medical image datasets across domains such as endoscopy, retinal fundus imaging, and facial acne photography. The experiments examine whether sparse relative comparisons are sufficient to learn a continuous severity score. Since the benchmark datasets provide absolute grades, we use the training labels only to construct pairwise relations in the form of more severe, less severe, or comparable. The absolute labels are retained only for the evaluation, and true class distribution is used for the known prior, which is reported separately from the label-free calibration routes.

\subsection{Datasets}

\textbf{LIMUC.}
The Labeled Images for Ulcerative Colitis (LIMUC)~\cite{polat2023improving} dataset is a publicly available colonoscopy dataset used to assess ulcerative colitis severity. It contains 11,276 images collected from 564 patients and 1,043 colonoscopy procedures. 

\textbf{IDRiD.}
The Indian Diabetic Retinopathy Image Dataset (IDRiD)~\cite{porwal2018indian} contains retinal fundus images labeled according to the diabetic retinopathy grading scale. The scale contains five ordinal levels, from grade 0 to grade 4.

\textbf{ACNE.}
ACNE04~\cite{wu2019joint} (simply referred to as ACNE) is a facial acne severity dataset containing 1,457 images with image-level ordinal acne grades. Although lesion-level bounding boxes are also provided, we use only the image-level severity grades, since our objective is ordinal severity estimation rather than lesion localization.

\subsection{Relative Annotation Protocol}
All three datasets provide absolute grades, which we use only to synthesize relative labels. For a pair budget of $p\%$, we sample $p\%$ of the $N(N-1)/2$ possible training pairs from the training set to form a candidate comparison pool and label each pair as $+1$, $-1$, or $0$ according to whether the first image has a higher, lower, or equal grade. The absolute grades are never given to the model as targets and they are used only to evaluate test predictions and, for the known-prior oracle, to compute the training class proportions. Each experiment is repeated with five different seeds for pair sampling and model training, and we report the mean and standard deviation over the three runs.

\subsection{Implementation Details}
The encoder uses an ImageNet-pretrained ConvNeXt-Tiny backbone followed by projection layers that produce a $128$-dimensional embedding. A lightweight scalar head maps this embedding to the continuous severity score. We set $\lambda_{\mathrm{rank}}=\lambda_{\mathrm{pp}}=0.5$, $\tau=0.07$, $L = 30$ and use a margin of $m=0.5$ for the pseudo-pair loss. The cumulative margin steps are parameterized using Softplus, initialized between $0.5$ and $2.0$ across the $K-1$ classes' ordinal distances. The pseudo-pairs are selected using $\tau_{\mathrm{pp}}=0.1$. During training, it uses resizing, horizontal flipping, rotation, and mild color jitter. All models are trained for $150$ epochs with a batch size of $32$, a learning rate of $10^{-4}$, weight decay of $10^{-4}$, and cosine learning-rate decay. For Tie-seg, we use $\lambda_{\mathrm{tie}}=1.0$ with a minimum bin size $1\%$ of the training samples.

\subsection{Evaluation Metrics}

We report accuracy and quadratic weighted kappa (QWK) since they capture complementary aspects of ordinal prediction. We also report the mean absolute error (MAE) in the supplementary material. Although accuracy measures exact agreement, it does not distinguish an adjacent-grade error from an error spanning several severity levels. Here, the QWK accounts for the ordered nature of the grades by assigning a larger penalty to predictions that are farther from the reference grade.

\newcommand{\ms}[2]{#1{\,\footnotesize$\pm$#2}}

\begin{table*}
\centering
\small
\setlength{\tabcolsep}{5.5pt}
\renewcommand{\ms}[2]{#1\,{\small$\pm$#2}}
\renewcommand{\arraystretch}{0.8}
\caption{\textbf{Main comparison on ConvNeXt-Tiny.}
Relative-supervision methods use sparse pairwise comparisons with Tie-seg calibration, while cross-entropy uses all absolute grades. Best relative-supervision results are in bold.}
\label{tab:main_comparison}
\begin{tabular}{@{}llc cc cc cc@{}}
\toprule
\multirow{2}{*}{Method} & \multirow{2}{*}{Supervision} & \multirow{2}{*}{Budget}
& \multicolumn{2}{c}{ACNE ($K{=}4$)}
& \multicolumn{2}{c}{IDRiD ($K{=}5$)}
& \multicolumn{2}{c}{LIMUC ($K{=}4$)} \\
\cmidrule(lr){4-5}\cmidrule(lr){6-7}\cmidrule(lr){8-9}
& & & Accuracy\ $\uparrow$ & QWK $\uparrow$ & Accuracy\ $\uparrow$ & QWK $\uparrow$
& Accuracy\ $\uparrow$ & QWK $\uparrow$ \\
\midrule
Cross-entropy & Absolute & 100\%
& \ms{78.6}{0.8} & \ms{0.822}{0.013}
& \ms{86.6}{0.4} & \ms{0.914}{0.008}
& \ms{78.0}{0.6} & \ms{0.848}{0.011} \\
\midrule
\multirow{5}{*}{MarginRank} & \multirow{5}{*}{Relative}
& 5\%  & \ms{68.6}{1.5} & \ms{0.743}{0.009} & \ms{73.6}{7.8} & \ms{0.812}{0.107} & \ms{55.7}{0.2} & \ms{0.252}{0.020} \\
& & 10\% & \ms{69.5}{3.4} & \ms{0.755}{0.043} & \textbf{\ms{80.5}{1.1}} & \ms{0.882}{0.008} & \ms{70.3}{1.1} & \ms{0.671}{0.046} \\
& & 30\% & \ms{71.2}{0.3} & \ms{0.788}{0.009} & \ms{81.2}{0.5} & \ms{0.878}{0.002} & \ms{73.9}{0.8} & \ms{0.844}{0.008} \\
& & 50\% & \ms{72.6}{2.4} & \ms{0.799}{0.020} & \ms{81.1}{0.1} & \ms{0.878}{0.004} & \ms{74.0}{0.6} & \ms{0.847}{0.005} \\
& & 70\% & \ms{73.1}{1.5} & \ms{0.786}{0.023} & \ms{81.9}{1.5} & \ms{0.890}{0.009} & \ms{76.1}{0.4} & \ms{0.839}{0.004} \\
\midrule
\multirow{5}{*}{RankNet} & \multirow{5}{*}{Relative}
& 5\%  & \ms{68.1}{1.2} & \ms{0.514}{0.009} & \ms{72.1}{9.1} & \textbf{\ms{0.841}{0.004}} & \textbf{\ms{69.4}{0.8}} & \textbf{\ms{0.599}{0.005}} \\
& & 10\% & \ms{66.7}{0.9} & \ms{0.487}{0.011} & \ms{72.3}{8.6} & \ms{0.841}{0.006} & \ms{71.0}{2.9} & \ms{0.674}{0.105} \\
& & 30\% & \ms{65.5}{1.7} & \ms{0.472}{0.022} & \ms{60.4}{0.6} & \ms{0.861}{0.004} & \ms{74.4}{0.4} & \ms{0.819}{0.003} \\
& & 50\% & \ms{66.0}{0.9} & \ms{0.487}{0.013} & \ms{61.0}{0.5} & \ms{0.864}{0.002} & \ms{74.4}{0.8} & \ms{0.819}{0.005} \\
& & 70\% & \ms{66.2}{2.2} & \ms{0.488}{0.029} & \ms{59.8}{0.0} & \ms{0.861}{0.001} & \ms{74.1}{0.6} & \ms{0.819}{0.004} \\
\midrule
\multirow{5}{*}{\textbf{DOSE}} & \multirow{5}{*}{Relative}
& 5\%  & \textbf{\ms{69.3}{1.0}} & \textbf{\ms{0.755}{0.009}} & \textbf{\ms{78.4}{0.4}} & \ms{0.839}{0.003} & \ms{56.4}{0.3} & \ms{0.498}{0.141} \\
& & 10\% & \textbf{\ms{70.2}{0.9}} & \textbf{\ms{0.762}{0.009}} & \ms{79.6}{0.8} & \textbf{\ms{0.900}{0.006}} & \textbf{\ms{75.2}{0.8}} & \textbf{\ms{0.821}{0.008}} \\
& & 30\% & \textbf{\ms{72.6}{1.2}} & \textbf{\ms{0.807}{0.011}} & \textbf{\ms{84.5}{0.7}} & \textbf{\ms{0.909}{0.010}} & \textbf{\ms{77.5}{0.4}} & \textbf{\ms{0.857}{0.003}} \\
& & 50\% & \textbf{\ms{74.8}{0.3}} & \textbf{\ms{0.821}{0.001}} & \textbf{\ms{84.0}{0.4}} & \textbf{\ms{0.903}{0.003}} & \textbf{\ms{78.0}{0.6}} & \textbf{\ms{0.862}{0.004}} \\
& & 70\% & \textbf{\ms{74.1}{2.0}} & \textbf{\ms{0.807}{0.021}} & \textbf{\ms{84.0}{0.4}} & \textbf{\ms{0.910}{0.008}} & \textbf{\ms{78.1}{0.4}} & \textbf{\ms{0.859}{0.004}} \\
\bottomrule
\end{tabular}
\end{table*}

\begin{table}[t!]
\centering
\scriptsize
\renewcommand{\ms}[2]{#1\,{\scriptsize$\pm$#2}}
\caption{
Calibration-route comparison using ConvNeXt-Tiny. Tie-seg, GMM, and Est-prior are label-free; Known-prior uses the true training class prior.
}
\label{tab:calibration_compact}

\begin{tabular}{@{}lc ccc c@{}}
\toprule
\multirow{2}{*}{Dataset}
& \multirow{2}{*}{Budget}
& \multicolumn{3}{c}{Label-free}
& \multicolumn{1}{c}{Oracle} \\
\cmidrule(lr){3-5}
\cmidrule(l){6-6}
&
& Tie-seg
& GMM
& Est-prior
& Known-prior \\
\midrule

\multirow{3}{*}{ACNE}
& 30\%
& \ms{72.6}{1.2}
& \textbf{\ms{73.6}{1.2}}
& \textbf{\ms{73.6}{0.0}}
& \ms{73.3}{0.9} \\

& 50\%
& \ms{74.8}{0.3}
& \ms{75.0}{1.0}
& \textbf{\ms{75.7}{0.6}}
& \ms{75.5}{0.9} \\

& 70\%
& \ms{74.1}{2.0}
& \textbf{\ms{75.1}{0.8}}
& \ms{74.4}{0.8}
& \ms{74.9}{1.3} \\

\midrule

\multirow{3}{*}{IDRiD}
& 30\%
& \textbf{\ms{84.5}{0.7}}
& \ms{84.2}{0.5}
& \ms{84.3}{1.1}
& \ms{84.7}{0.3} \\

& 50\%
& \textbf{\ms{84.0}{0.4}}
& \ms{83.6}{1.0}
& \ms{83.9}{0.7}
& \ms{84.7}{0.8} \\

& 70\%
& \textbf{\ms{84.0}{0.4}}
& \ms{83.8}{0.7}
& \ms{83.1}{0.7}
& \ms{84.8}{0.5} \\

\midrule

\multirow{3}{*}{LIMUC}
& 30\%
& \ms{77.5}{0.4}
& \textbf{\ms{77.6}{0.3}}
& \ms{76.2}{0.6}
& \ms{77.6}{0.5} \\

& 50\%
& \textbf{\ms{78.0}{0.6}}
& \textbf{\ms{78.0}{0.6}}
& \ms{76.6}{0.3}
& \ms{78.1}{0.6} \\

& 70\%
& \ms{78.1}{0.4}
& \textbf{\ms{78.2}{0.1}}
& \ms{77.3}{0.6}
& \ms{78.4}{0.2} \\

\bottomrule
\end{tabular}
\end{table}

\begin{table}[t]
\centering
\scriptsize
\setlength{\tabcolsep}{2.8pt}
\renewcommand{\arraystretch}{0.91}
\renewcommand{\ms}[2]{#1\,{\scriptsize$\pm$#2}}

\caption{Margin ablation with ConvNeXt-Tiny. Best results are in bold.}
\label{tab:ablation_margin}

\begin{tabular}{@{}lc cc cc cc@{}}
\toprule
\multirow{3}{*}{Dataset}
& \multirow{3}{*}{Budget}
& \multicolumn{4}{c}{Tie-seg}
& \multicolumn{2}{c}{Known-prior} \\
\cmidrule(lr){3-6}
\cmidrule(l){7-8}

&
& \multicolumn{2}{c}{Acc. $\uparrow$ (\%)}
& \multicolumn{2}{c}{QWK $\uparrow$}
& \multicolumn{2}{c}{Acc. $\uparrow$ (\%)} \\
\cmidrule(lr){3-4}
\cmidrule(lr){5-6}
\cmidrule(l){7-8}

&
& Multi & Single
& Multi & Single
& Multi & Single \\
\midrule

\multirow{3}{*}{ACNE}
& 30\%
& \textbf{\ms{72.6}{1.2}}
& \ms{69.4}{2.6}
& \textbf{0.807}
& 0.768
& \textbf{\ms{73.3}{0.9}}
& \ms{70.3}{1.7} \\

& 50\%
& \textbf{\ms{74.8}{0.3}}
& \ms{71.3}{2.5}
& \textbf{0.821}
& 0.785
& \textbf{\ms{75.5}{0.9}}
& \ms{71.1}{1.5} \\

& 70\%
& \textbf{\ms{74.1}{2.0}}
& \ms{71.6}{1.2}
& \textbf{0.807}
& 0.786
& \textbf{\ms{74.9}{1.3}}
& \ms{71.1}{1.7} \\

\midrule

\multirow{3}{*}{IDRiD}
& 30\%
& \textbf{\ms{84.5}{0.7}}
& \ms{82.8}{1.3}
& \textbf{0.909}
& 0.897
& \textbf{\ms{84.7}{0.3}}
& \ms{82.6}{0.8} \\

& 50\%
& \textbf{\ms{84.0}{0.4}}
& \ms{83.7}{1.0}
& \textbf{0.903}
& 0.902
& \textbf{\ms{84.7}{0.8}}
& \ms{82.5}{1.5} \\

& 70\%
& \ms{84.0}{0.4}
& \textbf{\ms{84.3}{1.0}}
& \textbf{0.910}
& 0.906
& \textbf{\ms{84.8}{0.5}}
& \ms{83.1}{1.0} \\

\midrule

\multirow{3}{*}{LIMUC}
& 30\%
& \textbf{\ms{77.5}{0.4}}
& \ms{76.9}{0.3}
& \textbf{0.857}
& 0.840
& \textbf{\ms{77.6}{0.5}}
& \ms{77.1}{0.4} \\

& 50\%
& \textbf{\ms{78.0}{0.6}}
& \ms{75.7}{0.1}
& \textbf{0.862}
& 0.826
& \textbf{\ms{78.1}{0.6}}
& \ms{76.0}{0.0} \\

& 70\%
& \textbf{\ms{78.1}{0.4}}
& \ms{76.5}{0.7}
& \textbf{0.859}
& 0.835
& \textbf{\ms{78.4}{0.2}}
& \ms{77.1}{0.4} \\

\bottomrule
\end{tabular}
\end{table}
\noindent

\section{Results}
\subsection{Comparison with Fully Supervised and Pairwise-Ranking Baselines}

Table~\ref{tab:main_comparison} compares DOSE with a fully supervised cross-entropy classifier using the same ConvNeXt-Tiny backbone and a linear layer on top of the backbone. The supervised model uses every absolute label, whereas DOSE is trained only on sparse relative comparisons. As expected, full supervision yields higher accuracy on ACNE and IDRiD, with gains of $ 3.8\%$ and $2.1\%$ over the strongest DOSE result, respectively. However, DOSE remains close in QWK, differing only by $0.001$ on ACNE and $0.004$ on IDRiD. On LIMUC, DOSE matches or slightly exceeds the supervised classifier, reaching $78.1\%$ accuracy and a QWK of $0.862$. These results show that sparse relative supervision can retain much of the performance obtained with complete absolute labels supervision.

We further compare DOSE with MarginRank and RankNet using the ConvNeXt-Tiny backbone and the label-free Tie-seg route. These methods use the same relative comparisons and pair budgets; therefore, the comparison isolates the effect of the proposed dominance-guided objective. DOSE achieves the highest accuracy and QWK across almost all evaluated settings, with a few exceptions. The improvement is particularly clear on IDRiD, where DOSE reaches $84.0$--$84.5\%$ accuracy, compared with $81.1$--$81.9\%$ for MarginRank and approximately $60\%$ for RankNet. Also, the consistent gains on ACNE and LIMUC further show that DOSE improves both accuracy and ordinal agreement.

\subsection{Comparison of Calibration Routes}

Table~\ref{tab:calibration_compact} compares the four score-to-grade routes using DOSE with ConvNeXt-Tiny. The label-free routes perform similarly on ACNE and LIMUC, while Tie-seg is the most stable route on IDRiD across all pair budgets. Tie-seg also remains close to the known prior, showing that the learned severity axis can be converted into discrete grades without access to the true class distribution.

\subsection{Ablation of the Margin Formulation}
\label{sec:ablation}















\begin{figure}
    \centering
    \includegraphics[width=\columnwidth]{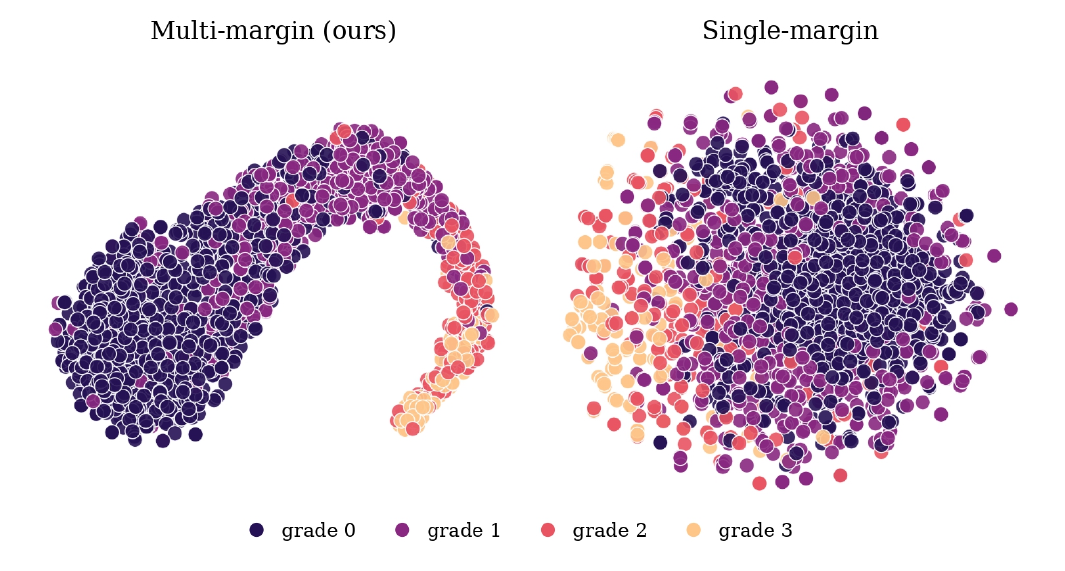}
\caption{Comparison of the embedding. Left: our multi-margin loss. Right: single-margin variant. This comparison uses the LIMUC dataset with a 30\% pair budget.}
    \label{fig:qualitative}
\end{figure}

Table~\ref{tab:ablation_margin} compares the cumulative multi-margin objective with a single shared margin using ConvNeXt-Tiny. All other training and calibration components are unchanged. Multi-margin improves Tie-seg accuracy in eight of nine settings, with average gains of $3.1$, $0.6$, and $1.5$ \% points on ACNE, IDRiD, and LIMUC, respectively. It also improves QWK, reduces MAE, and achieves higher
known-prior accuracy across every pair budget. These results indicate that distance-dependent margins yield a more clearly separated ordinal representation of severity. Fig.~\ref{fig:qualitative} shows the qualitative behavior of our multi-margin loss on the embedding $\mathbf{z}$ when compared with the single-margin-style loss. Complete results across backbones, including QWK and MAE, are provided in the supplementary material.


\noindent

\section{Conclusion}

We proposed a novel framework for estimating ordinal medical severity grades using pairwise comparisons rather than relying on absolute grades. The comparison graph is used in our method to compute global dominance scores, which are then converted into ordinal distance levels. We also make use of dominance-based pseudo-pairs to provide additional supervision for a continuous-severity head. The discrete grades are then extracted from the learned severity axis through three different label-free calibration routes and a known-prior oracle. We also created a multi-margin using relative labels, which outperforms a single margin across almost all settings, demonstrating that images with larger estimated severity differences should receive greater separation. The Tie-seg route remains close to the known prior across all backbones and pair budgets (see supplementary). In case of an imbalanced score distribution, it is more stable than density-based calibration. However, the relative labels are synthesized from absolute labels, so they contain no real annotator disagreement, and at a very low pair budget, the comparison graph becomes too sparse for reliable dominance estimates. Our future work will be to evaluate DOSE on real expert comparisons with disagreement, and to reduce the number of required comparisons through active pair selection and model-comparison uncertainty.


\section*{Acknowledgments}
This work was supported in part by the Japan Society for the Promotion of Science (JSPS) KAKENHI under Grant Number JP24K03010.

{
    \small\bibliographystyle{ieeenat_fullname}
    \bibliography{main}
}
\clearpage
\appendix

 
 
 
 
 
 
\setcounter{table}{0}
\renewcommand{\thetable}{S\arabic{table}}
\setcounter{figure}{0}
\renewcommand{\thefigure}{S\arabic{figure}}




\section{Additional Experimental Results}

This supplementary material provides the complete results omitted from the main paper due to space constraints. We report calibration results and the ordinal-margin ablation across ConvNeXt-Tiny, ResNet-50, and DenseNet-121, including Accuracy, QWK, and MAE.

\subsection{Calibration Results Across Backbones}

Table~\ref{tab:backbone_full} extends the calibration comparison in the main paper to all three backbones. Tie-seg, GMM, and Est-prior remain label-free, while Known-prior uses the true training class prior as an oracle reference.

\begin{table*}[t]
\centering
\footnotesize
\setlength{\tabcolsep}{2.5pt}
\caption{Label-free calibration across backbones and pair budgets, with Accuracy, QWK, and
MAE for every route. \emph{Tie-seg} (prior-free),
\emph{GMM}, and \emph{Est-prior} use no labels; \emph{Known-prior} uses the true class
prior. Per row, the best label-free accuracy is in bold.}
\label{tab:backbone_full}
\resizebox{1.0\textwidth}{!}{
\begin{tabular}{llc ccc ccc ccc ccc}
\toprule
\multirow{2}{*}{Dataset} & \multirow{2}{*}{Backbone} & \multirow{2}{*}{Budget}
 & \multicolumn{3}{c}{Tie-seg} & \multicolumn{3}{c}{GMM}
 & \multicolumn{3}{c}{Est-prior} & \multicolumn{3}{c}{Known-prior} \\
\cmidrule(lr){4-6}\cmidrule(lr){7-9}\cmidrule(lr){10-12}\cmidrule(lr){13-15}
 & & & Acc $\uparrow$ & QWK $\uparrow$ & MAE$\downarrow$ & Acc $\uparrow$ & QWK $\uparrow$ & MAE$\downarrow$ & Acc $\uparrow$ & QWK $\uparrow$  & MAE$\downarrow$ & Acc $\uparrow$ & QWK $\uparrow$ & MAE$\downarrow$ \\
\midrule
\multirow{9}{*}{ACNE ($K{=}4$)}
 & \multirow{3}{*}{ConvNeXt-T}
   & 30\% & \ms{72.6}{1.2} & \ms{0.807}{0.011} & \ms{0.274}{0.012} & \textbf{\ms{73.6}{1.2}} & \ms{0.821}{0.012} & \ms{0.264}{0.012} & \textbf{\ms{73.6}{0.0}} & \ms{0.818}{0.003} & \ms{0.264}{0.000} & \ms{73.3}{0.9} & \ms{0.819}{0.010} & \ms{0.267}{0.009} \\
 & & 50\% & \ms{74.8}{0.3} & \ms{0.821}{0.001} & \ms{0.252}{0.003} & \ms{75.0}{1.0} & \ms{0.824}{0.007} & \ms{0.252}{0.009} & \textbf{\ms{75.7}{0.6}} & \ms{0.818}{0.003} & \ms{0.248}{0.003} & \ms{75.5}{0.9} & \ms{0.826}{0.008} & \ms{0.248}{0.009} \\
 & & 70\% & \ms{74.1}{2.0} & \ms{0.807}{0.021} & \ms{0.263}{0.023} & \textbf{\ms{75.1}{0.8}} & \ms{0.825}{0.005} & \ms{0.251}{0.008} & \ms{74.4}{0.8} & \ms{0.813}{0.007} & \ms{0.260}{0.010} & \ms{74.9}{1.3} & \ms{0.820}{0.014} & \ms{0.256}{0.016} \\
\cmidrule(l){2-15}
 & \multirow{3}{*}{ResNet-50}
   & 30\% & \ms{72.7}{1.8} & \ms{0.785}{0.020} & \ms{0.280}{0.020} & \textbf{\ms{74.0}{2.0}} & \ms{0.810}{0.018} & \ms{0.266}{0.020} & \ms{73.1}{0.9} & \ms{0.793}{0.007} & \ms{0.276}{0.006} & \ms{73.7}{1.1} & \ms{0.798}{0.012} & \ms{0.271}{0.012} \\
 & & 50\% & \ms{73.9}{1.7} & \ms{0.804}{0.012} & \ms{0.266}{0.017} & \textbf{\ms{74.3}{1.1}} & \ms{0.812}{0.007} & \ms{0.264}{0.011} & \ms{74.0}{1.6} & \ms{0.799}{0.015} & \ms{0.266}{0.014} & \ms{75.0}{0.9} & \ms{0.812}{0.011} & \ms{0.257}{0.010} \\
 & & 70\% & \ms{72.1}{0.8} & \ms{0.783}{0.010} & \ms{0.284}{0.003} & \textbf{\ms{73.3}{1.5}} & \ms{0.803}{0.005} & \ms{0.273}{0.010} & \ms{70.0}{2.5} & \ms{0.750}{0.030} & \ms{0.309}{0.021} & \ms{73.4}{1.4} & \ms{0.795}{0.012} & \ms{0.274}{0.010} \\
\cmidrule(l){2-15}
 & \multirow{3}{*}{DenseNet-121}
   & 30\% & \ms{73.6}{1.9} & \ms{0.805}{0.014} & \ms{0.269}{0.017} & \textbf{\ms{74.1}{2.6}} & \ms{0.813}{0.016} & \ms{0.264}{0.024} & \ms{72.9}{2.0} & \ms{0.789}{0.021} & \ms{0.279}{0.017} & \ms{73.7}{2.1} & \ms{0.802}{0.013} & \ms{0.271}{0.017} \\
 & & 50\% & \ms{73.4}{1.2} & \ms{0.801}{0.006} & \ms{0.271}{0.010} & \ms{73.1}{1.4} & \ms{0.809}{0.013} & \ms{0.274}{0.016} & \textbf{\ms{74.0}{1.2}} & \ms{0.796}{0.017} & \ms{0.270}{0.014} & \ms{72.7}{0.8} & \ms{0.801}{0.012} & \ms{0.280}{0.012} \\
 & & 70\% & \textbf{\ms{74.1}{1.4}} & \ms{0.799}{0.017} & \ms{0.266}{0.017} & \ms{73.7}{1.6} & \ms{0.808}{0.019} & \ms{0.269}{0.018} & \ms{72.9}{3.0} & \ms{0.780}{0.025} & \ms{0.283}{0.028} & \ms{73.6}{1.9} & \ms{0.799}{0.016} & \ms{0.273}{0.019} \\
\midrule
\multirow{9}{*}{IDRiD ($K{=}5$)}
 & \multirow{3}{*}{ConvNeXt-T}
   & 30\% & \textbf{\ms{84.5}{0.7}} & \ms{0.909}{0.010} & \ms{0.196}{0.013} & \ms{84.2}{0.5} & \ms{0.924}{0.005} & \ms{0.186}{0.006} & \ms{84.3}{1.1} & \ms{0.925}{0.008} & \ms{0.184}{0.014} & \ms{84.7}{0.3} & \ms{0.916}{0.003} & \ms{0.190}{0.004} \\
 & & 50\% & \textbf{\ms{84.0}{0.4}} & \ms{0.903}{0.003} & \ms{0.205}{0.005} & \ms{83.6}{1.0} & \ms{0.922}{0.005} & \ms{0.193}{0.010} & \ms{83.9}{0.7} & \ms{0.921}{0.004} & \ms{0.191}{0.008} & \ms{84.7}{0.8} & \ms{0.919}{0.005} & \ms{0.187}{0.008} \\
 & & 70\% & \textbf{\ms{84.0}{0.4}} & \ms{0.910}{0.008} & \ms{0.198}{0.008} & \ms{83.8}{0.7} & \ms{0.923}{0.006} & \ms{0.191}{0.010} & \ms{83.1}{0.7} & \ms{0.920}{0.007} & \ms{0.198}{0.011} & \ms{84.8}{0.5} & \ms{0.918}{0.002} & \ms{0.190}{0.005} \\
\cmidrule(l){2-15}
 & \multirow{3}{*}{ResNet-50}
   & 30\% & \textbf{\ms{82.2}{0.9}} & \ms{0.877}{0.007} & \ms{0.239}{0.011} & \ms{36.3}{2.5} & \ms{0.790}{0.006} & \ms{0.707}{0.020} & \ms{31.9}{5.7} & \ms{0.771}{0.018} & \ms{0.768}{0.056} & \ms{82.5}{0.2} & \ms{0.900}{0.007} & \ms{0.220}{0.007} \\
 & & 50\% & \textbf{\ms{82.8}{0.6}} & \ms{0.888}{0.011} & \ms{0.227}{0.011} & \ms{34.1}{1.8} & \ms{0.783}{0.006} & \ms{0.731}{0.018} & \ms{29.3}{4.0} & \ms{0.763}{0.013} & \ms{0.801}{0.041} & \ms{83.1}{0.7} & \ms{0.900}{0.006} & \ms{0.218}{0.010} \\
 & & 70\% & \textbf{\ms{82.5}{0.6}} & \ms{0.886}{0.003} & \ms{0.230}{0.006} & \ms{32.5}{1.9} & \ms{0.779}{0.010} & \ms{0.745}{0.023} & \ms{24.2}{4.5} & \ms{0.745}{0.013} & \ms{0.852}{0.044} & \ms{83.6}{0.3} & \ms{0.903}{0.007} & \ms{0.211}{0.007} \\
\cmidrule(l){2-15}
 & \multirow{3}{*}{DenseNet-121}
   & 30\% & \textbf{\ms{83.2}{0.9}} & \ms{0.902}{0.008} & \ms{0.214}{0.011} & \ms{46.7}{17.0} & \ms{0.815}{0.043} & \ms{0.599}{0.182} & \ms{45.9}{17.9} & \ms{0.809}{0.048} & \ms{0.619}{0.199} & \ms{83.1}{1.0} & \ms{0.905}{0.004} & \ms{0.212}{0.011} \\
 & & 50\% & \textbf{\ms{83.1}{1.0}} & \ms{0.890}{0.009} & \ms{0.223}{0.013} & \ms{56.1}{20.1} & \ms{0.844}{0.054} & \ms{0.495}{0.221} & \ms{54.6}{20.9} & \ms{0.834}{0.054} & \ms{0.521}{0.232} & \ms{83.6}{1.0} & \ms{0.904}{0.006} & \ms{0.210}{0.011} \\
 & & 70\% & \textbf{\ms{83.1}{0.9}} & \ms{0.891}{0.003} & \ms{0.221}{0.007} & \ms{39.2}{2.1} & \ms{0.799}{0.007} & \ms{0.683}{0.019} & \ms{37.2}{1.5} & \ms{0.784}{0.011} & \ms{0.723}{0.024} & \ms{83.8}{0.9} & \ms{0.909}{0.005} & \ms{0.205}{0.010} \\
\midrule
\multirow{9}{*}{LIMUC ($K{=}4$)}
 & \multirow{3}{*}{ConvNeXt-T}
   & 30\% & \ms{77.5}{0.4} & \ms{0.857}{0.003} & \ms{0.228}{0.003} & \textbf{\ms{77.6}{0.3}} & \ms{0.860}{0.001} & \ms{0.228}{0.002} & \ms{76.2}{0.6} & \ms{0.851}{0.003} & \ms{0.241}{0.006} & \ms{77.6}{0.5} & \ms{0.860}{0.003} & \ms{0.228}{0.004} \\
 & & 50\% & \textbf{\ms{78.0}{0.6}} & \ms{0.862}{0.004} & \ms{0.223}{0.006} & \textbf{\ms{78.0}{0.6}} & \ms{0.864}{0.004} & \ms{0.223}{0.007} & \ms{76.6}{0.3} & \ms{0.857}{0.003} & \ms{0.236}{0.004} & \ms{78.1}{0.6} & \ms{0.866}{0.005} & \ms{0.222}{0.007} \\
 & & 70\% & \ms{78.1}{0.4} & \ms{0.859}{0.004} & \ms{0.223}{0.004} & \textbf{\ms{78.2}{0.1}} & \ms{0.863}{0.003} & \ms{0.222}{0.002} & \ms{77.3}{0.6} & \ms{0.861}{0.004} & \ms{0.231}{0.007} & \ms{78.4}{0.2} & \ms{0.865}{0.003} & \ms{0.220}{0.002} \\
\cmidrule(l){2-15}
 & \multirow{3}{*}{ResNet-50}
   & 30\% & \textbf{\ms{76.7}{0.7}} & \ms{0.841}{0.004} & \ms{0.240}{0.007} & \textbf{\ms{76.7}{0.8}} & \ms{0.847}{0.004} & \ms{0.239}{0.008} & \ms{76.3}{0.7} & \ms{0.846}{0.004} & \ms{0.242}{0.006} & \ms{76.8}{0.7} & \ms{0.845}{0.002} & \ms{0.239}{0.007} \\
 & & 50\% & \ms{76.5}{0.5} & \ms{0.845}{0.004} & \ms{0.240}{0.005} & \textbf{\ms{76.6}{0.6}} & \ms{0.851}{0.004} & \ms{0.239}{0.007} & \ms{75.8}{0.4} & \ms{0.845}{0.004} & \ms{0.247}{0.004} & \ms{76.7}{0.6} & \ms{0.851}{0.004} & \ms{0.239}{0.006} \\
 & & 70\% & \ms{76.7}{0.7} & \ms{0.843}{0.008} & \ms{0.239}{0.008} & \textbf{\ms{76.8}{0.6}} & \ms{0.848}{0.005} & \ms{0.238}{0.006} & \ms{76.1}{0.8} & \ms{0.847}{0.005} & \ms{0.245}{0.007} & \ms{77.0}{0.7} & \ms{0.850}{0.007} & \ms{0.236}{0.007} \\
\cmidrule(l){2-15}
 & \multirow{3}{*}{DenseNet-121}
   & 30\% & \ms{76.0}{0.5} & \ms{0.839}{0.003} & \ms{0.245}{0.004} & \textbf{\ms{76.3}{0.6}} & \ms{0.846}{0.003} & \ms{0.242}{0.006} & \ms{75.2}{1.1} & \ms{0.841}{0.008} & \ms{0.252}{0.010} & \ms{76.6}{0.5} & \ms{0.847}{0.004} & \ms{0.239}{0.004} \\
 & & 50\% & \textbf{\ms{76.4}{0.5}} & \ms{0.844}{0.003} & \ms{0.241}{0.004} & \ms{76.3}{0.8} & \ms{0.849}{0.004} & \ms{0.241}{0.007} & \ms{75.8}{0.9} & \ms{0.849}{0.006} & \ms{0.247}{0.009} & \ms{76.4}{0.7} & \ms{0.850}{0.004} & \ms{0.241}{0.007} \\
 & & 70\% & \ms{76.3}{0.3} & \ms{0.844}{0.004} & \ms{0.242}{0.004} & \textbf{\ms{76.6}{0.4}} & \ms{0.851}{0.004} & \ms{0.239}{0.005} & \ms{75.7}{0.6} & \ms{0.848}{0.003} & \ms{0.248}{0.006} & \ms{76.5}{0.3} & \ms{0.851}{0.003} & \ms{0.239}{0.003} \\

\bottomrule
\end{tabular}
}
\end{table*}

\subsection{Margin Ablation Across Backbones}

Table~\ref{tab:ablation_margin} extends the margin ablation in the main
paper to ResNet-50 and DenseNet-121 and additionally reports MAE.
All other components are kept unchanged between the multi-margin and
single-margin variants.

\begin{table*}[t]
\centering \footnotesize
\setlength{\tabcolsep}{3.5pt}
\caption{
\textbf{Ablation of the ordinal margin formulation.}
The proposed cumulative multi-margin objective is compared with a single learnable margin under identical pair budgets, backbones, dominance estimation, and calibration protocols.}
\label{tab:ablation_margin}
\begin{tabular}{llc cc cc cc cc}
\toprule
\multirow{2}{*}{Dataset} & \multirow{2}{*}{Backbone} & \multirow{2}{*}{Budget}
 & \multicolumn{6}{c}{Tie-seg } & \multicolumn{2}{c}{Known-prior} \\
\cmidrule(lr){4-9}\cmidrule(lr){10-11}
 & & & \multicolumn{2}{c}{Accuracy $\uparrow$ (\%)} & \multicolumn{2}{c}{QWK $\uparrow$}
     & \multicolumn{2}{c}{MAE $\downarrow$} & \multicolumn{2}{c}{Accuracy $\uparrow$ (\%)} \\
\cmidrule(lr){4-5}\cmidrule(lr){6-7}\cmidrule(lr){8-9}\cmidrule(lr){10-11}
 & & & Multi & Single & Multi & Single & Multi & Single & Multi & Single \\
\midrule
\multirow{9}{*}{ACNE ($K{=}4$)}
 & \multirow{3}{*}{ConvNeXt-T}
   & 30\% & \textbf{\ms{72.6}{1.2}} & \ms{69.4}{2.6} & \textbf{0.807} & 0.768 & \textbf{0.274} & 0.309 & \textbf{\ms{73.3}{0.9}} & \ms{70.3}{1.7} \\
 & & 50\% & \textbf{\ms{74.8}{0.3}} & \ms{71.3}{2.5} & \textbf{0.821} & 0.785 & \textbf{0.252} & 0.294 & \textbf{\ms{75.5}{0.9}} & \ms{71.1}{1.5} \\
 & & 70\% & \textbf{\ms{74.1}{2.0}} & \ms{71.6}{1.2} & \textbf{0.807} & 0.786 & \textbf{0.263} & 0.289 & \textbf{\ms{74.9}{1.3}} & \ms{71.1}{1.7} \\
\cmidrule(l){2-11}
 & \multirow{3}{*}{ResNet-50}
   & 30\% & \textbf{\ms{72.7}{1.8}} & \ms{68.7}{2.3} & \textbf{0.785} & 0.742 & \textbf{0.280} & 0.317 & \textbf{\ms{73.7}{1.1}} & \ms{70.4}{1.6} \\
 & & 50\% & \textbf{\ms{73.9}{1.7}} & \ms{67.3}{0.8} & \textbf{0.804} & 0.716 & \textbf{0.266} & 0.339 & \textbf{\ms{75.0}{0.9}} & \ms{69.1}{0.9} \\
 & & 70\% & \textbf{\ms{72.3}{1.7}} & \ms{67.1}{1.6} & \textbf{0.783} & 0.744 & \textbf{0.286} & 0.331 & \textbf{\ms{73.6}{2.1}} & \ms{69.0}{1.9} \\
\cmidrule(l){2-11}
 & \multirow{3}{*}{DenseNet-121}
   & 30\% & \textbf{\ms{73.6}{1.9}} & \ms{69.4}{1.5} & \textbf{0.805} & 0.748 & \textbf{0.269} & 0.313 & \textbf{\ms{73.7}{2.1}} & \ms{69.3}{1.7} \\
 & & 50\% & \textbf{\ms{73.4}{1.2}} & \ms{69.3}{2.2} & \textbf{0.801} & 0.761 & \textbf{0.271} & 0.311 & \textbf{\ms{72.7}{0.8}} & \ms{71.0}{2.7} \\
 & & 70\% & \textbf{\ms{74.1}{1.4}} & \ms{68.1}{2.8} & \textbf{0.799} & 0.738 & \textbf{0.266} & 0.327 & \textbf{\ms{73.6}{1.9}} & \ms{68.7}{2.6} \\
\midrule
\multirow{9}{*}{IDRiD ($K{=}5$)}
 & \multirow{3}{*}{ConvNeXt-T}
   & 30\% & \textbf{\ms{84.5}{0.7}} & \ms{82.8}{1.3} & \textbf{0.909} & 0.897 & \textbf{0.196} & 0.221 & \textbf{\ms{84.7}{0.3}} & \ms{82.6}{0.8} \\
 & & 50\% & \textbf{\ms{84.0}{0.4}} & \ms{83.7}{1.0} & \textbf{0.903} & 0.902 & \textbf{0.205} & 0.213 & \textbf{\ms{84.7}{0.8}} & \ms{82.5}{1.5} \\
 & & 70\% & \ms{84.0}{0.4} & \textbf{\ms{84.3}{1.0}} & \textbf{0.910} & 0.906 & \textbf{0.198} & 0.203 & \textbf{\ms{84.8}{0.5}} & \ms{83.1}{1.0} \\
\cmidrule(l){2-11}
 & \multirow{3}{*}{ResNet-50}
   & 30\% & \textbf{\ms{82.2}{0.9}} & \ms{81.8}{1.1} & 0.877 & \textbf{0.886} & 0.239 & \textbf{0.234} & \textbf{\ms{82.5}{0.2}} & \ms{82.2}{1.1} \\
 & & 50\% & \textbf{\ms{82.8}{0.6}} & \ms{81.5}{1.1} & \textbf{0.888} & 0.884 & \textbf{0.227} & 0.239 & \textbf{\ms{83.1}{0.7}} & \ms{82.6}{0.6} \\
 & & 70\% & \textbf{\ms{82.5}{0.6}} & \ms{80.1}{0.9} & \textbf{0.886} & 0.877 & \textbf{0.230} & 0.254 & \textbf{\ms{83.6}{0.3}} & \ms{81.5}{0.5} \\
\cmidrule(l){2-11}
 & \multirow{3}{*}{DenseNet-121}
   & 30\% & \textbf{\ms{83.2}{0.9}} & \ms{81.5}{1.6} & \textbf{0.902} & 0.890 & \textbf{0.214} & 0.234 & \textbf{\ms{83.1}{1.0}} & \ms{80.4}{1.0} \\
 & & 50\% & \textbf{\ms{83.1}{1.0}} & \textbf{\ms{83.1}{2.4}} & 0.890 & \textbf{0.897} & 0.223 & \textbf{0.221} & \textbf{\ms{83.6}{1.0}} & \ms{81.6}{1.0} \\
 & & 70\% & \textbf{\ms{83.1}{0.9}} & \ms{81.9}{1.3} & 0.891 & \textbf{0.894} & \textbf{0.221} & 0.231 & \textbf{\ms{83.8}{0.9}} & \ms{82.2}{1.1} \\
\midrule
\multirow{9}{*}{LIMUC ($K{=}4$)}
 & \multirow{3}{*}{ConvNeXt-T}
   & 30\% & \textbf{\ms{77.5}{0.4}} & \ms{76.9}{0.3} & \textbf{0.857} & 0.840 & \textbf{0.228} & 0.239 & \textbf{\ms{77.6}{0.5}} & \ms{77.1}{0.4} \\
 & & 50\% & \textbf{\ms{78.0}{0.6}} & \ms{75.7}{0.1} & \textbf{0.862} & 0.826 & \textbf{0.223} & 0.251 & \textbf{\ms{78.1}{0.6}} & \ms{76.0}{0.0} \\
 & & 70\% & \textbf{\ms{78.1}{0.4}} & \ms{76.5}{0.7} & \textbf{0.859} & 0.835 & \textbf{0.223} & 0.242 & \textbf{\ms{78.4}{0.2}} & \ms{77.1}{0.4} \\
\cmidrule(l){2-11}
 & \multirow{3}{*}{ResNet-50}
   & 30\% & \textbf{\ms{76.7}{0.7}} & \ms{75.3}{0.8} & \textbf{0.841} & 0.833 & \textbf{0.240} & 0.253 & \textbf{\ms{76.8}{0.7}} & \ms{75.7}{0.4} \\
 & & 50\% & \textbf{\ms{76.5}{0.5}} & \ms{76.2}{0.5} & \textbf{0.845} & 0.838 & \textbf{0.240} & 0.245 & \textbf{\ms{76.7}{0.6}} & \ms{76.2}{0.4} \\
 & & 70\% & \textbf{\ms{76.7}{0.7}} & \ms{75.9}{0.5} & \textbf{0.843} & 0.837 & \textbf{0.239} & 0.249 & \textbf{\ms{77.0}{0.7}} & \ms{76.0}{0.6} \\
\cmidrule(l){2-11}
 & \multirow{3}{*}{DenseNet-121}
   & 30\% & \textbf{\ms{76.0}{0.5}} & \ms{75.2}{0.7} & \textbf{0.839} & 0.825 & \textbf{0.245} & 0.257 & \textbf{\ms{76.6}{0.5}} & \ms{75.4}{0.8} \\
 & & 50\% & \textbf{\ms{76.4}{0.5}} & \ms{75.6}{0.8} & \textbf{0.844} & 0.831 & \textbf{0.241} & 0.250 & \textbf{\ms{76.4}{0.7}} & \ms{75.8}{1.0} \\
 & & 70\% & \textbf{\ms{76.3}{0.3}} & \ms{75.1}{0.7} & \textbf{0.844} & 0.830 & \textbf{0.242} & 0.256 & \textbf{\ms{76.5}{0.3}} & \ms{75.6}{0.6} \\
\bottomrule
\end{tabular}
\end{table*}

\section{Additional Details of Dominance Estimation}

We provide the details of the random-walk transition matrix used for global dominance estimation in Sec.~3.3 of the main paper.

Let $A_{uv}$ denote the number of observed comparisons in which image $u$ is judged more severe than image $v$. Tied comparisons are excluded. The maximum degree of
the non-tied comparison graph is defined as
\begin{equation}
d_{\max} =
\max_u \sum_{v\neq u}
\mathbb{I}[A_{uv}+A_{vu}>0].
\end{equation}

Following Rank Centrality [28], we construct the
transition matrix $P$ as
\begin{equation}
P_{uv} =
\begin{cases}
\displaystyle
\frac{A_{vu}}{d_{\max}(A_{uv}+A_{vu})},
& u\neq v,\ A_{uv}+A_{vu}>0,\\[6pt]
\displaystyle
1-\sum_{w\neq u}P_{uw},
& u=v,\\[6pt]
0, & \text{otherwise}.
\end{cases}
\end{equation}

The off-diagonal transitions move toward images judged more severe, while the diagonal entries provide self-loops to ensure that each row sums to one. Unobserved and tied pairs, therefore, contribute no off-diagonal transitions.

Starting from the uniform distribution $\boldsymbol{\pi}^{(0)}=\mathbf{1}/N$, we perform $L$ power iterations:
\begin{equation}
\boldsymbol{\pi}^{(t+1)}
=P^\top\boldsymbol{\pi}^{(t)},
\qquad t=0,\ldots,L-1.
\end{equation}
The probability vector is renormalized after each iteration for numerical stability. We use $\boldsymbol{\pi}=\boldsymbol{\pi}^{(L)}$ to derive the dominance scores. Thus, convergence to a unique stationary distribution is not required.


\end{document}